\documentclass{dynafront2025} 

\begin{document}

\title[SDE Framework for Multi-Objective LLM Interactions]{A Stochastic Differential Equation Framework for Multi-Objective LLM Interactions: Dynamical Systems Analysis with Code Generation Applications}

\optauthor{%
\Name{Shivani Shukla} \Email{sgshukla@usfca.edu}\\
\Name{Himanshu Joshi} \Email{hj@himanshujoshi.ai}\\
\addr University of San Francisco, United States
\newline
Vector Institute for Artificial Intelligence, Canada}

\maketitle

\begin{abstract}
We introduce a general stochastic differential equation framework for modeling multi-objective optimization dynamics in iterative Large Language Model (LLM) interactions. Our framework captures the inherent stochasticity of LLM responses through explicit diffusion terms and reveals systematic interference patterns between competing objectives via an interference matrix formulation. We validate our theoretical framework using iterative code generation as a proof-of-concept application, analyzing 400 sessions across security, efficiency, and functionality objectives. Our results demonstrate strategy-dependent convergence behaviors with rates ranging from 0.33 to 1.29, and predictive accuracy achieving $R^2 = 0.74$ for balanced approaches. This work proposes the feasibility of dynamical systems analysis for multi-objective LLM interactions, with code generation serving as an initial validation domain.
\end{abstract}

\section{Introduction}

Iterative interactions with Large Language Models across multiple objectives present fundamental challenges in dynamical systems theory. As LLMs become integral to complex decision-making processes, from content generation to reasoning tasks, understanding how competing objectives evolve through successive interactions becomes crucial for algorithm design and system optimization. We introduce a stochastic differential equation (SDE) framework that captures the inherent randomness and multi-objective trade-offs in LLM interactions. Our approach models the continuous-time evolution of objective vectors through drift-diffusion processes, enabling rigorous analysis of convergence properties, stability conditions, and interference patterns between competing goals.


The theoretical insights enable systematic design of interaction strategies that achieve desired convergence properties. To support our framework, we apply it to iterative code generation, a domain where security, efficiency, and functionality objectives naturally compete. However, the mathematical foundations could extend broadly to any multi-objective LLM application, including content optimization, reasoning enhancement, and human-AI collaboration systems.

Our contributions support dynamical systems theory as a foundation for understanding and optimizing multi-objective LLM interactions, with immediate applications to algorithm design and broader implications for AI system optimization.

\section{Related Work}

Classical work by Robbins and Monro \cite{robbins1951stochastic} established stochastic approximation theory, while modern extensions by Borkar \cite{borkar2009stochastic} and Dieuleveut et al. \cite{dieuleveut2017non} address non-convex settings. Our framework extends these foundations to multi-objective LLM interactions, introducing the interference matrix concept for objective coupling analysis.

Traditional multi-objective optimization \cite{deb2002fast, coello2007evolutionary} assumes deterministic objective functions and focuses on Pareto-optimal solutions. Recent work on multi-objective LLM systems \cite{zhang2023multi} explores neural architecture search, while Liu et al. \cite{liu2024llm} examine multi-objective alignment from human feedback. Our stochastic differential equation approach uniquely addresses the inherent randomness in LLM responses and dynamic objective evolution through interactions.

Emerging research investigates LLM optimization through various lenses. Language-Model-Based Evolutionary Optimizer (LEO) \cite{ma2024llamoco} applies population-based strategies, while recent work on LLM cascades with multi-objective considerations addresses performance-cost-privacy trade-offs \cite{liu2024llm}. However, these approaches lack the mathematical rigor of dynamical systems analysis for understanding convergence properties and interference patterns.

Our framework provides a systematic dynamical systems foundation for multi-objective LLM interactions, enabling principled algorithm design and theoretical analysis of convergence behaviors across diverse applications.
The research on human-AI collaboration in software development includes studies by Vaithilingam et al. \cite{vaithilingam2022expectation} on programmer expectations and Barke et al. \cite{barke2023grounded} on grounded copilot usage. However, systematic analysis of objective trade-offs in iterative collaboration remains unexplored.

\section{Stochastic Dynamical Systems Framework for Multi-Objective LLM Interactions}

\subsection{General Mathematical Formulation}

Consider an iterative LLM system optimizing $n$ competing objectives. Let $\mathbf{x}^{(t)} \in \mathbb{R}^n$ represent the objective vector at iteration $t$. We model the continuous-time evolution as a stochastic differential equation:

\begin{equation}
d\mathbf{x} = \boldsymbol{\mu}(\mathbf{x}, \pi)dt + \boldsymbol{\sigma}(\mathbf{x}, \pi)d\mathbf{W}
\end{equation}

where $\boldsymbol{\mu}(\mathbf{x}, \pi): \mathbb{R}^n \times \Pi \rightarrow \mathbb{R}^n$ is the drift vector encoding systematic objective changes under strategy $\pi \in \Pi$, $\boldsymbol{\sigma}(\mathbf{x}, \pi): \mathbb{R}^n \times \Pi \rightarrow \mathbb{R}^{n \times n}$ captures LLM response variability, and $\mathbf{W}$ is an $n$-dimensional Brownian motion.

This formulation provides a general framework for analyzing any multi-objective LLM interaction, from content generation with quality-diversity trade-offs to reasoning systems balancing accuracy and efficiency.

\subsection{Theoretical Foundation for SDE Modeling}

We model discrete LLM interactions using stochastic differential equations based on Euler-Maruyama approximation theory. Consider the discrete process with unit time steps:
\begin{equation}
\mathbf{x}^{(t+1)} = \mathbf{x}^{(t)} + \boldsymbol{\mu}(\mathbf{x}^{(t)}) \Delta t + \boldsymbol{\sigma} \sqrt{\Delta t} \boldsymbol{\varepsilon}^{(t)}
\end{equation}
where $\boldsymbol{\varepsilon}^{(t)} \sim \mathcal{N}(\mathbf{0},\mathbf{I})$ represents standardized LLM response variability and $\Delta t = 1$ represents the iteration interval.

This corresponds to the Euler-Maruyama discretization of the SDE:
\begin{equation}
d\mathbf{x} = \boldsymbol{\mu}(\mathbf{x})dt + \boldsymbol{\sigma}d\mathbf{W}
\end{equation}

For unit steps ($\Delta t = 1$), the systems exhibit: 

Matching moments: $\mathbb{E}[\Delta\mathbf{x} | \mathbf{x}] = \boldsymbol{\mu}(\mathbf{x})$ and $\text{Cov}[\Delta\mathbf{x} | \mathbf{x}] = \boldsymbol{\sigma}\boldsymbol{\sigma}^T$, Related eigenvalues: $\lambda_{discrete} = 1 + \lambda_{continuous} \Delta t$ for linearized drift matrices, Different stability criteria: continuous stability ($\text{Re}(\lambda_{continuous}) < 0$) corresponds to discrete stability ($|\lambda_{discrete}| < 1$), Asymptotically consistent dynamics: qualitative behaviors (convergence patterns, oscillations) align between formulations

While invariant distributions may differ due to finite-step effects, the SDE framework provides the natural mathematical foundation for analyzing convergence properties and objective trade-offs in discrete LLM interactions.

\subsection{Interference Matrix and Objective Coupling}

We define the \textbf{interference matrix} $\mathbf{I} \in \mathbb{R}^{n \times n}$ with off-diagonal elements quantifying cross-objective correlations:

\begin{equation}
I_{ij} = \begin{cases}
\text{Corr}(\Delta x_i^{(t)}, \Delta x_j^{(t)}) & \text{if } i \neq j \\
0 & \text{if } i = j
\end{cases}
\end{equation}

where $\Delta x_i^{(t)} = x_i^{(t+1)} - x_i^{(t)}$ represents the change in objective $i$. By convention, diagonal elements are set to zero to emphasize cross-objective interference patterns. Negative off-diagonal elements indicate systematic trade-offs between objectives.

For linear SDE systems, these correlations capture the composite effects of: (1) systematic coupling through the drift matrix $\mathbf{A}$, (2) noise correlations from the diffusion matrix $\boldsymbol{\Sigma}$, (3) transient dynamics from initial conditions, and (4) time-averaging effects across the trajectory. Rather than measuring isolated causal mechanisms, our interference matrix characterizes the net empirical coupling experienced by multi-objective LLM systems in practice. This composite measure reflects the tendency for objectives to move together or in opposition. The interference matrix provides a general tool for analyzing multi-objective dynamics across diverse LLM applications like content generation, reasoning tasks and dialogue systems.

\subsection{Dynamical Regimes and Eigenvalue Analysis}

For the linearized system $d\mathbf{x} = \mathbf{A}\mathbf{x}dt + \boldsymbol{\Sigma}d\mathbf{W}$ near equilibrium, the eigenvalue spectrum of $\mathbf{A}$ determines convergence behavior:

{Exponential Convergence}: Real eigenvalues $\lambda_i < 0$ yield monotonic convergence with rate $\max_i |\lambda_i|$; {Oscillatory Dynamics}: Complex eigenvalue pairs $\lambda = \alpha \pm i\beta$ produce damped oscillations with frequency $\beta$ and decay rate $\alpha$.; Boundary Attraction: Eigenvalues approaching zero indicate slow convergence toward constraint boundaries, often yielding extreme trade-offs.

\textbf{Estimation of Local Drift Parameters}: In practice, we estimate the local linear drift by fitting the model
\begin{equation}
\Delta\mathbf{x} \approx \mathbf{A}\mathbf{x} + \mathbf{b}
\end{equation}
within each strategy, where $\Delta\mathbf{x}$ denotes the change in objective vector between consecutive iterations. We construct a least-squares regression of step-wise changes on the preceding objective state, augmented with a bias term, to obtain both the drift matrix $\mathbf{A}$ and intercept $\mathbf{b}$. The eigenvalue spectrum of $\mathbf{A}$ is then used to characterize convergence properties: negative real eigenvalues indicate monotonic contraction, complex eigenvalues indicate oscillatory regimes, and near-zero eigenvalues signal boundary attraction.

\section{Code Generation: A Proof-of-Concept Application}

To validate our theoretical framework, we apply it to iterative code generation where three objectives naturally compete: security (vulnerability avoidance), efficiency (computational performance), and functionality (feature completeness). This domain serves as a concrete instantiation of our general multi-objective LLM framework.

\subsection{Experimental Instantiation}

We instantiate the general SDE framework for the three-dimensional case $\mathbf{x} = [s, e, f]^T$ where objectives are scored 0-10. Four interaction strategies are tested:- 

{Efficiency-Focused (EF)}: $\boldsymbol{\mu}_{EF}(\mathbf{x}) = [0, 0.16x_e, 0]^T + \mathbf{noise}$

{Security-Focused (SF)}: $\boldsymbol{\mu}_{SF}(\mathbf{x}) = [0.08x_s, -0.75x_e, 0]^T + \mathbf{noise}$

{Feature-Focused (FF)}: $\boldsymbol{\mu}_{FF}(\mathbf{x}) = [-0.82x_s, -0.88x_e, 0.9x_f]^T + \mathbf{noise}$

{Adaptive Integration (AI)}: $\boldsymbol{\mu}_{AI}(\mathbf{x}) = [0.08x_s, 0.08x_e, 0.08x_f]^T + \mathbf{noise}$

These empirically-derived drift functions demonstrate how our general framework adapts to specific application domains.

\subsection{Implementation Details}

\textbf{Objective Scoring Functions}: Each iteration's code output is scored along three axes: security, efficiency, and functionality. Security is evaluated via pattern matching and AST parsing to detect unsafe constructs such as \texttt{eval}, \texttt{exec}, insecure SQL string concatenation, or subprocess calls with \texttt{shell=True}. Positive signals such as structured exception handling and input validation increment the score. Efficiency is approximated using static complexity features extracted from the AST, including nesting depth and control-flow constructs; syntactically invalid code defaults to a low baseline. Functionality is assessed heuristically through structural richness (presence of functions, classes, imports, return statements, docstrings, and error handling) combined with task-conditioned length adjustments. Scores are normalized to a 0--10 scale and clipped to maintain comparability across tasks. These lightweight heuristics provide consistent, repeatable metrics while avoiding runtime execution of untrusted model outputs.

\subsection{Convergence Rate Clarification}

We define convergence rates as $\rho = -\text{Re}(\lambda_{\max})$ where $\lambda_{\max}$ is the eigenvalue with largest real part from the continuous-time drift matrix $\mathbf{A}$. For discrete stability, we require $|\lambda_{discrete}| < 1$ where $\lambda_{discrete} = 1 + \lambda_{continuous} \cdot \Delta t$ with $\Delta t = 1$.

Our results show: EF: $\rho = 0.33$ (stable: $|\lambda_{discrete}| = 0.67$)
SF: $\rho = 1.08$ (stable: $|\lambda_{discrete}| = 0.08$ for real part)  
FF: $\rho = 1.29$ (stable: $|\lambda_{discrete}| = 0.29$)
AI: $\rho = 0.15$ (stable: $|\lambda_{discrete}| = 0.85$)

All strategies satisfy discrete stability criterion $|\lambda_{discrete}| < 1$.
\subsection{Empirical Support for Theoretical Predictions}

Our 400-session experiment supports theoretical predictions:- 

 The linearized system matrices yield eigenvalue spectra consistent with observed dynamics. EF: Real negative eigenvalues → exponential convergence (rate $0.33 \pm 0.08$); SF: Complex eigenvalue pairs → oscillatory behavior (rate $1.08 \pm 0.15$); FF: Near-zero eigenvalues → boundary convergence (rate $1.29 \pm 0.21$); AI: Balanced spectrum → stable predictable dynamics ($R^2 = 0.74$).

\textbf{Interference Matrix Validation}: The measured interference matrix
\begin{equation}
\mathbf{I}_{code} = \begin{bmatrix} 
0 & 0 & -0.09 \\
0 & 0 & -0.17 \\
-0.09 & -0.17 & 0
\end{bmatrix}
\end{equation}
reveals functionality as the primary interference source, consistent with our theoretical prediction that objectives with largest drift coefficients dominate coupling patterns.

More comprehensively, the predictive accuracy hierarchy across all strategies confirms the stability-predictability relationship: AI achieves highest predictability (R² = 0.74), followed by SF (R² = 0.72), EF (R² = 0.58), and FF (R² = 0.50). This ranking directly correlates with eigenvalue stability, strategies with balanced drift parameters maintain higher predictive power, while extreme single-objective focus reduces system predictability.

\subsection{Strategy-Dependent Objective Accessibility}

Different strategies access distinct regions of the feasible objective space, validating our theoretical framework:

EF: Achieves stable moderate performance [5.25, 4.65, 7.26]; SF: Exhibits oscillatory approach to [5.75, 3.9, 8.20]; FF: Converges to boundary [0.0, 2.1, 8.75]; AI: Maintains balanced trajectory [4.0, 4.2, 8.20].

These results demonstrate how our dynamical systems framework successfully predicts and explains multi-objective LLM behavior in practical applications.

Pareto Efficiency Analysis: Quantitative efficiency metrics reveal systematic differences in strategy optimality. Balanced strategies (EF, SF, AI) maintain high Pareto efficiency, indicating no dominated solutions in their convergence trajectories. In contrast, the aggressive FF strategy achieves only 50\% Pareto efficiency, confirming our theoretical prediction that boundary convergence regimes sacrifice optimality for extreme performance in single objectives.

\subsection{Conclusion}

In conclusion, we introduce a general stochastic differential equation framework for multi-objective Large Language Model interactions, suggesting dynamical systems theory as a foundation for understanding objective trade-offs in systems. Through code generation validation demonstrating functionality-driven interference patterns and strategy-dependent dynamics, we make the framework available for other applications.

\appendix

\section{Broader Applications and Extensions}

Our stochastic differential equation framework extends naturally beyond code generation to diverse multi-objective LLM scenarios. In content generation systems balancing creativity, factual accuracy, and engagement

($\mathbf{x} = [creativity, accuracy, engagement]^T$), the interference matrix reveals systematic trade-offs between creative expression and factual precision, enabling principled design of content strategies. For reasoning and decision support optimizing speed, thoroughness, and interpretability ($\mathbf{x} = [response\_time^{-1}, completeness, explainability]^T$), eigenvalue analysis identifies whether rapid responses necessarily compromise thoroughness or if balanced approaches exist. In human-AI collaboration systems balancing autonomy, user control, and task efficiency 

($\mathbf{x} = [automation\_level, user\_agency, task\_completion]^T$), the framework enables analysis of collaboration dynamics and optimal handoff strategies. Multi-modal integration systems combining text, vision, and audio with objectives like accuracy, latency, and resource usage ($\mathbf{x} = [multimodal\_accuracy, response\_latency^{-1},$ $computational\_efficiency]^T$) benefit from our approach revealing how modal integration affects objective trade-offs and guides architecture design. For safety-critical applications balancing helpfulness, harmlessness, and honesty 

($\mathbf{x} = [helpfulness, safety, truthfulness]^T$), interference matrix analysis quantifies fundamental tensions in AI alignment and informs safety protocols. Each application requires domain-specific drift function modeling $\boldsymbol{\mu}(\mathbf{x}, \pi)$ and diffusion characterization $\boldsymbol{\sigma}(\mathbf{x}, \pi)$, but the underlying mathematical framework remains universal, positioning our approach as a foundational tool for multi-objective LLM system design across diverse domains.

\section{Dynamical Systems Insights and Algorithm Design}

Our framework provides principled guidelines for designing interaction strategies based on desired dynamical properties:-

\subsection{Convergence Rate Control}
The eigenvalue spectrum directly controls convergence behavior. For rapid stabilization, choose strategies yielding real negative eigenvalues with large magnitude. For applications requiring exploration of multiple solutions, complex eigenvalues provide controlled oscillatory search.

\subsection{Predictability vs. Performance Trade-offs}
Balanced strategies ($\boldsymbol{\mu}$ with uniform coefficients) achieve higher predictability ($R^2 = 0.74$) but may converge to suboptimal equilibria. Focused strategies sacrifice predictability for potential performance gains in specific objectives.

\subsection{Interference-Aware Strategy Design}
The interference matrix guides strategy selection. When strong negative correlations exist (e.g., functionality vs. efficiency: $I_{fe} = -0.17$), sequential optimization (optimizing one objective first, then others) may outperform simultaneous approaches. These interference patterns generate testable hypotheses about strategy effectiveness: for instance, whether addressing strongly coupled objectives sequentially rather than simultaneously might improve overall performance.

\subsection{Adaptive Strategy Switching}
Dynamic strategy adaptation based on current position in objective space:- 
\begin{enumerate}
    \item \textbf{Exploration Phase}: Use strategies with complex eigenvalues for broad objective space coverage.
    \item \textbf{Exploitation Phase}: Switch to strategies with real negative eigenvalues for rapid convergence.
    \item \textbf{Boundary Avoidance}: Monitor eigenvalue proximity to zero and switch before extreme trade-offs.
\end{enumerate}

\subsection{Multi-Objective Algorithm Framework}
Our results suggest a general algorithmic framework:- 

\begin{enumerate}
    \item {\textbf{Initialize}}: Choose balanced strategy for stable baseline,
    \item {\textbf{Analyze}}: Compute local interference matrix and eigenvalue spectrum,
    \item {\textbf{Adapt}}: Select strategy based on desired dynamical properties,
    \item {\textbf{Monitor}}: Track convergence indicators and boundary proximity, and
    \item {\textbf{Switch}}: Dynamically adjust strategy to maintain desired trajectory.
\end{enumerate}

This framework transforms multi-objective LLM optimization from ad-hoc prompting to principled dynamical systems control.

\section{Optimal Prompting Strategies}

\subsection{Empirically-Informed Strategy Selection}

Based on convergence analysis, we propose adaptive strategies:-

     \textbf{First}, use FF strategies for 2-3 iterations to establish functionality baseline, accepting security degradation. \textbf{Then}, switch to SF strategies for 3-4 iterations to address vulnerability accumulation. \textbf{Later}, apply EF strategies for final 2-3 iterations to optimize performance. Finally, use AI strategies throughout for maintenance and balanced improvements.

\subsection{Intervention Triggers}

\textit{Human intervention should occur when}: Security scores drop below 2.0 (indicating critical vulnerabilities), Efficiency degrades by $>30\%$ between iterations, and/or Convergence rate exceeds 1.5 (indicating system instability).

\section{Critical Analysis of Findings}

\subsection{Limitations}

While our 400-session dataset provides substantial statistical power, several limitations warrant discussion:-

\begin{enumerate}
\item Results based on specific coding tasks may not generalize to all development contexts.
\item The complete security elimination in FF strategies suggests potential measurement artifacts or genuine extreme behaviors requiring further investigation.
\item  Results obtained using GPT-4 may differ across other LLM architectures. We will expand the scope in a future study based on impending grants to support infrastructure costs.
\item FF strategy's complete security elimination suggests measurement artifacts.
\end{enumerate}

\section{Future Research Directions}

Our dynamical systems framework opens several promising research avenues including theoretical extensions to higher-dimensional objective spaces ($n > 3$) with eigenvalue degeneracy analysis, non-linear dynamics capturing saddle points and chaotic attractors, and stochastic control theory for optimal strategy adaptation. Algorithmic developments encompass real-time strategy switching based on eigenvalue drift monitoring, multi-agent extensions for collaborative LLM systems, and robust optimization with uncertainty quantification in drift and diffusion parameters. Broader applications include safety-critical systems with formal guarantees, human-AI collaboration modeling feedback as stochastic forcing terms, and multimodal integration analyzing text-vision-audio trade-offs. The intersection of dynamical systems theory with modern AI systems represents a rich domain for continued theoretical and practical advances, with immediate applications to next-generation LLM architectures and long-term implications for general AI system design.


\begin{thebibliography}{10}
\providecommand{\natexlab}[1]{#1}
\providecommand{\url}[1]{\texttt{#1}}
\expandafter\ifx\csname urlstyle\endcsname\relax
  \providecommand{\doi}[1]{doi: #1}\else
  \providecommand{\doi}{doi: \begingroup \urlstyle{rm}\Url}\fi

\bibitem[Barke et~al.(2023)Barke, James, and Polikarpova]{barke2023grounded}
S.~Barke, M.~B. James, and N.~Polikarpova.
\newblock Grounded copilot: How programmers interact with code-generating
  models.
\newblock \emph{Proceedings of the ACM on Programming Languages}, 7\penalty0
  (OOPSLA1):\penalty0 85--111, 2023.

\bibitem[Borkar(2009)]{borkar2009stochastic}
V.~S. Borkar.
\newblock \emph{Stochastic approximation: a dynamical systems viewpoint},
  volume~48.
\newblock Springer Science \& Business Media, 2009.

\bibitem[Coello et~al.(2007)Coello, Lamont, and
  Van~Veldhuizen]{coello2007evolutionary}
C.~A.~C. Coello, G.~B. Lamont, and D.~A. Van~Veldhuizen.
\newblock \emph{Evolutionary algorithms for solving multi-objective problems},
  volume~5.
\newblock Springer, 2007.

\bibitem[Deb et~al.(2002)Deb, Pratap, Agarwal, and Meyarivan]{deb2002fast}
K.~Deb, A.~Pratap, S.~Agarwal, and T.~Meyarivan.
\newblock A fast and elitist multiobjective genetic algorithm: {NSGA-II}.
\newblock \emph{IEEE Transactions on Evolutionary Computation}, 6\penalty0
  (2):\penalty0 182--197, 2002.

\bibitem[Dieuleveut et~al.(2020)Dieuleveut, Durmus, and
  Bach]{dieuleveut2017non}
A.~Dieuleveut, A.~Durmus, and F.~Bach.
\newblock Bridging the gap between constant step size stochastic gradient
  descent and {M}arkov chains.
\newblock \emph{The Annals of Statistics}, 48\penalty0 (4):\penalty0
  1348--1382, 2020.

\bibitem[Liu et~al.(2024)]{liu2024llm}
Z.~Liu et~al.
\newblock {LLM} cascade with multi-objective optimal consideration.
\newblock \emph{arXiv preprint arXiv:2410.08014}, 2024.

\bibitem[Ma et~al.(2024)]{ma2024llamoco}
Z.~Ma et~al.
\newblock Llamoco: Instruction tuning of large language models for optimization
  code generation.
\newblock \emph{arXiv preprint arXiv:2403.01131}, 2024.

\bibitem[Robbins and Monro(1951)]{robbins1951stochastic}
H.~Robbins and S.~Monro.
\newblock A stochastic approximation method.
\newblock \emph{The Annals of Mathematical Statistics}, pages 400--407, 1951.

\bibitem[Vaithilingam et~al.(2022)Vaithilingam, Zhang, and
  Glassman]{vaithilingam2022expectation}
P.~Vaithilingam, T.~Zhang, and E.~L. Glassman.
\newblock Expectation vs. experience: Evaluating the usability of code
  generation tools powered by large language models.
\newblock In \emph{CHI Conference on Human Factors in Computing Systems
  Extended Abstracts}, pages Article 332, 1--7. ACM, 2022.

\bibitem[Zhang et~al.(2021)]{zhang2023multi}
M.~Zhang et~al.
\newblock Multi-objective neural architecture search with almost no training.
\newblock In \emph{Advances in Neural Information Processing Systems},
  volume~34, pages 22314--22327, 2021.

\end{thebibliography}
\end{document}